\documentclass[journal]{IEEEtran}
\usepackage{widetext}

\usepackage{booktabs}
\usepackage{multirow}
\usepackage{xcolor}
\usepackage[normalem]{ulem}

\ifCLASSINFOpdf
  \usepackage[pdftex]{graphicx}
  \graphicspath{{../pdf/}{../jpeg/}}
  \DeclareGraphicsExtensions{.pdf,.jpeg,.png}
\else
  \usepackage[dvips]{graphicx}
  \graphicspath{{../eps/}}
  \DeclareGraphicsExtensions{.eps}
\fi
\usepackage{xcolor}

\usepackage{amsmath, amssymb, amsfonts}
\usepackage{amsfonts}
\usepackage{bm}
\usepackage{url}

\usepackage{subcaption}

\begin{document}
%
\title{VMAF Re-implementation on PyTorch: \\ Some Experimental Results  \\
\vspace {0.5cm}
\large{\bf Huawei Technical Report \\}
\large{\bf Cloud BU \\}
}

%
%
%
\author{
    \IEEEauthorblockN{Kirill Aistov\IEEEauthorrefmark{1}, Maxim Koroteev\IEEEauthorrefmark{1}}
    \\\IEEEauthorblockA{\IEEEauthorrefmark{1} \small Huawei Russian Research Institute, Moscow
    \\}

\thanks{Moscow Research Center, Moscow, Russia. Cloud BU.  E-mail: koroteev.maxim@huawei.com; kirill.aistov1@huawei.com}}
\maketitle

\begin{abstract}
Based on the standard VMAF implementation we propose an implementation of VMAF using PyTorch framework. For this implementation comparisons with the standard (libvmaf) show negligible discrepancy in VMAF units. We investigate gradients computation when using VMAF as an objective function and demonstrate that training using this function does not result in ill-behaving gradients. The implementation is then validated by using VMAF as a loss function to train a preprocessing filter. It is demonstrated that its performance is superior to the unsharp masking filter. The resulting filter is also easy for implementation and can be applied in video processing tasks for video compression improvement. This is confirmed by the results of numerical experiments.
\end{abstract}

%
\IEEEpeerreviewmaketitle


\section*{Introduction}

Video Multimethod Assessment Fusion (VMAF) developed by Netflix \cite{VMAF} was released in 2016 and quickly gained popularity due to its high correlation with subjective quality metrics. It has become in recent years one of the main tools used for image/video quality assessment for compression tasks in both research and industry. In the same time it was shown that VMAF score can be significantly increased by certain preprocessing methods, e.g., sharpening or histogram equalization \cite{siniukov2021hacking}; this led Netflix to release an alternative version of the metric referred to as VMAF NEG that is less susceptible to such preprocessing\footnote{It is worth noting that such preprocessing may not necessarily be harmful: its goal can be improving the image quality.}. It seems natural that the further application of this metric may require some automatic ML (machine learning) methods to be trained using such a powerful tool as VMAF. The original VMAF algorithm was implemented in C \cite{vmaf_github} and no effort is known to us to re-implement it {\it fully, i.e., including all its sub-metrics} using some ML framework. One of the reasons for that may be the claimed non-differentiability of this metric. 

It is known that the frameworks such as PyTorch and Tensorflow allow to compute gradients of functions via automatic differentiation \cite{autodiff}. Without the use of automatic differentiation the code to produce gradients has to be implemented manually\footnote{libvmaf does not provide this code.} or the gradients have to be computed using approximate formulas, however this is very inefficient in the case of a large number of parameters. 
The basic library {\it libvmaf}, containing VMAF, does not provide an efficient tool to compute VMAF gradients.
This prevented the direct use of VMAF as an objective function for learning ML models. We propose an implementation of VMAF using PyTorch and analyze its differentiability with various methods. Thus, one of the issues we tried to resolve in this paper is possibility of the direct use of the PyTorch implementation of VMAF for ML.

Using this approach we then study how this implementation can be combined with the stochastic gradient descent to learn a preprocessing filter. The results demonstrate significant improvement over such a standard preprocessing method as unsharp masking. We also discuss potential problems related to the computation of VMAF metric in the end of the paper.

\section*{Construction of VMAF}

First, we provide a short review of VMAF structure. VMAF score is computed by calculating two elementary image metrics referred to as VIF and ADM (sometimes DLM) for each frame, and a so-called "Motion" feature; the final score is produced via SVM regression that uses these features as an input. Here we provide brief descriptions for these features.

\subsection{VIF} 

VIF (visual information fidelity)\cite{VIF} computes a ratio of two mutual information measures between images under the assumptions of the gaussian channel model for image distortion and HVS (human visual system). Roughly speaking the algorithm can be described as follows. For the reference image patches$\{C_i\}_{i=1}^{N}$ and distorted image patches $\{D_i\}_{i=1}^{N}$ one computes the ratio of two mutual informations which has the form
\begin{equation}
	VIF=\frac{\sum_{i=1}^{N}\log_2 \left(1+\frac{g_i^2 \cdot \sigma_{C_i}^2}{\sigma_{V_i}^2+\sigma_N^2}\right)}{\sum_{i=1}^{N}\log_2 \left(1+\frac{\sigma_{C_i}^2}{\sigma_N^2}\right)},
	\label{vif}
\end{equation}
where parameters in (\ref{vif}) are those of the gaussian channel models (not written down here explicitly, see \cite{VIF} for details) and $g_{i}$ and $\sigma_{V_i}^2$ are estimated as
$$
g_{i}=\frac{\sigma_{C_iD_i}}{\sigma_{C_i}^2 }, \\         
$$
$$
\sigma_{V_i}^2=\sigma_{D_i}^2-g_i \cdot \sigma_{C_iD_i}, \\ 
$$
and $\sigma_N^2$ is a variance of the gaussian noise incorporated into the HVS model. Note, that these estimates in principle have to be computed over the sample of images. Instead, the assumption is made that the estimates can be computed over the patches (\cite{VIF}, section IV; \cite{VIF2})

VIF is computed on four scales by downsampling the image; four values per frame are used as features for final score regression. 
The original version of VIF included the wavelet transform, but the same authors released another version of VIF in the pixel domain \cite{VIF_pixel_domain}. VMAF uses only the pixel domain version, so it is this version we implemented in our work\footnote{The PyTorch implementation of the wavelet domain version is also available and can be found at \url{https://github.com/chaofengc/IQA-PyTorch/blob/main/pyiqa/archs/vif\_arch.py}}.

\subsection{ADM} 

ADM (Additive Detail Metric) \cite{ADM} operates in the wavelet domain, the metric tries to decompose the target (distorted) image $T$ into a restored imaged $R$ using the original image $O$ and an additive impairment image $A$: $T=R+A$ where 
\begin{equation}
R = 
\left\{
\begin{array}{cr}
	\operatorname{clip}_{[0,1]}\left(\frac{T}{O}\right)O , & \text{if } \left|\Psi_O-\Psi_T\right|>1^{\circ}\\
	T, & \text{if }\left|\Psi_O-\Psi_T\right|\le1^{\circ}.
\end{array}
\right.
\end{equation}
Here $\Psi$ is arctan of the ratio between two coefficients co-located in the vertical subband and the horizontal subband of the same scale, the special case $\left|\Psi_O-\Psi_T\right|<1^{\circ}$ is made to handle contrast enhancement. For more information refer to the original paper \cite{ADM} or \cite{VMAFNEG}.

After decoupling the original image $O$ goes through contrast sensitivity function (CSF) and the restored image $R$ goes through a contrast sensitivity function and a contrast masking (CM) function. CSF is computed by multiplying wavelet coefficients of each subband with its corresponding CSF value. CM function is computed by convolving these coefficients with a specific kernel and a thresholding operation. 

The final score is computed using the formula:
$$
A D M=\frac{\sum_{\lambda=1}^4 \sum_{\theta=2}^4 \left(\sum_{i, j \in \text { C }}{\rm CM}({\rm CSF}(R(\lambda, \theta, i, j)))^3\right)^{\frac{1}{3}}}
{\sum_{\lambda=1}^4 \sum_{\theta=2}^4 \left(\sum_{i, j \in \text { C }} {\rm CSF}(O(\lambda, \theta, i, j))^3\right)^{\frac{1}{3}}},
$$
where $R(\lambda, \theta, i, j)$ and $O(\lambda, \theta, i, j)$ are wavelet coefficients of the restored and original image at scale $\lambda$, subband $\theta$ (vertical $\theta=2$, horizontal $\theta=4$ and diagonal $\theta=3$) and spatial coefficients $i,j$, $C$ represents the central area of the image (coefficients at the outer edge are ignored).

The default VMAF version uses a single value from ADM per frame for the final score regression. Alternatively, four values for four scales from ADM can be computed by omitting the first sum in the formula and used as individual features.

\subsection{Motion} 

The motion feature for frame $i$ is computed using the formula 
$$
\min({\rm SAD}(f_i,f_{i-1}) ,{\rm SAD}(f_i,f_{i+1}) ),
$$
where $f_i$ is frame $i$ after smoothing using a $5\times5$ gaussian filter, and SAD is the sum of absolute differences between pixels.
This is the only feature that contains temporal information.

\subsection*{Regression } 

The features described above can be computed for each frame of a video stream; all features use only the luma component of the frame. A score for each frame is produced using SVM regression (after feature normalizaton). SVM uses an RBF kernel; given a feature vector $x$, the score is computed with the following formula 
$$
\sum_{i \in SV}\alpha_i K(x_i, x) + b, K(u,v)=\exp(-\gamma ||u-v||^2),
$$ 
where $x_i$ are support vectors.
The final score for the video is produced by taking the average of frame scores and clipping it to $[0, 100]$ range.

\subsection*{VMAF NEG} 

VMAF NEG version modifies the formulas used to calculate VIF and ADM elementary features by introducing parameters called enhancement gain limit (EGL) and modifying (essentially clipping) certain internal values based on these parameters: for VIF 
\begin{equation}
g_i = \min(g_i, EGL_{V I F})
\label{vmaf_neg_clipping}
\end{equation}
for ADM
$$ 
R=\min \left(R \cdot E G L_{D L M}, T\right) \text{, if }\left|\Psi_O-\Psi_T\right|<1^{\circ} \text { and } R>0,
$$
$$ 
R=\max \left(R \cdot E G L_{D L M}, T\right) \text{, if }\left|\Psi_O-\Psi_T\right|<1^{\circ} \text { and } R<0. 
$$

For a more detailed description and reasoning behind this see \cite{VMAFNEG} and \cite{VMAF2}. 

\section*{Numerical experiments} 

We implement both the base VMAF algorithm and NEG version in PyTorch framework. This is to our knowledge the first implementation to allow gradient based optimization. We closely follow the official Netflix implementation in C \cite{vmaf_github} in order to obtain output values as close as possible to it. The difference in scores measured over $79$ video streams provided by Netflix public dataset\cite{netflix} is $\le 0.01\pm 0.01$ VMAF units (using first $50$ frames from each video); note, VMAF scales in the interval $0-100$ and for typical natural images VMAF takes on values around $80-95$ so the error is by order $10^{-4}$ smaller than actual VMAF values measured for natural images. We also compare all elementary features for two implementations. It was found that the difference is from $\approx 7\times 10^{-6}$ for ADM to $\approx 2\times 10^{-4}$ for Motion on the same data. So it seems the latter metric is least precisely reproduced even though the numbers show that this precision is sufficient for the majority of applications. The small differences observed for sub-metrics probably occur because of discrepancies in image padding which are different in PyTorch and the official implementation in libvmaf; this issue will be investigated further. Some small difference is also likely due to the fact that default libvmaf version uses quantized integer values for performance reasons and our PyTorch version uses floating point values to allow differentiation. 

\section*{Gradient checking}

VIF, ADM and motion features along with the final score regression are mostly composed of simple tensor manipulations, convolution operations (for downsampling, wavelet transform and contrast masking), and elementary functions such as exponents and logarithms, which are differentiable. The problem to computing gradients may emerge from  operations such as clipping and ReLU which produce gradients equal to zero in some part of their domain. We observe that gradients computed in the case of default VMAF version do not approach to machine precision zero, e.g., $\sim 10^{-16}$.
Another peculiarity is the fact that ADM as implemented in VMAF uses only central area of the image and ignores the outer edge, so the ADM gradients for outer edge pixels are zero. However this is compensated by VIF gradients.

To ensure that gradients are computed correctly we perform a procedure known as gradient checking (see e.g., \cite{gradient_checking}). 
Given some function $f(\theta)$ and a function $g(\theta)$ that is supposed to compute $\frac{\partial f}{\partial \theta}$ we can ensure that $g(\theta)$ is correct by numerically verifying
$$
g(\theta) \approx \frac{f(\theta+\varepsilon)-f(\theta-\varepsilon)}{2\varepsilon}
$$

In the case of VMAF gradient checking is complicated by the fact that reference C implementation takes files in .yuv format as input i.e the input values can be only integer numbers in $[0,255]$.
To perform gradient checking we compute the derivative of a very simple learnable image transform -- a convolution with a single filter kernel. We perform this on single frame.
If $R$ is a reference image, $W=\{W_{ij}\}_{i,j=1}^{k}$ is the convolution kernel, $R \ast W$ is the output of the convolution, we compute 
$$
\frac{\partial \operatorname{VMAF(R, R \ast W)}}{\partial W_{ij}}
$$
by backpropagation algorithm using the PyTorch version.

Let matrices $W^{(km+)}$, $W^{(km-)}$ be defined by
$$
W^{(km\pm)}_{ij} = W_{ij} \pm \varepsilon\delta_{ki}\delta_{mj},
$$
where $\delta_{ki}$, $\delta_{mj}$ are Kronecker deltas. Then we compute the central difference approximation of the derivative as
$$
\frac{\operatorname{VMAF(R, R \ast W^{(km+)})}-\operatorname{VMAF(R, R \ast W^{(km-)})}}{2\varepsilon}
$$
using the reference C version.
We round the output of the convolutions $R \ast W^{(km+)}$ and $R \ast W^{(km-)}$ to nearest integer before giving it as input to VMAF C version.

Initialization for the filter weights should be done carefully since we need all pixels of the resulting image to be in $[0,255]$ range. We initialize each element with $\frac{1}{k^2}$, where $k$ is the size of the filter to ensure that the average brightness does not change. The tests were performed with the filters of sizes $k=3$ and $k=5$.

It is clear that the finite difference approximation of the derivative becomes inexact when $\varepsilon$ grows so this parameter can not be made too big. On the other hand, in the case of small $\varepsilon$ the outputs of the perturbed convolutions $R \ast W^{(km+)}$ and $R \ast W^{(km-)}$ may differ by the magnitude smaller than one pixel and if the differences are $<0.5$, then rounding will remove the impact of perturbation. The output of the perturbed convolution should of course also be in $[0,255]$. Taking all this into account we set $\varepsilon=10^{-2}$.

We find that in the settings described above the derivatives are close (taking into account that rounding introduces additional error): for the central coefficient of $3\times3$ filter the derivative computed numerically using C implementation is $223.8$ and the derivative computed by means of backpropagation using PyTorch is $223.4$. We compared derivatives for all elements of the kernel and found that average difference is $0.41\pm 0.35$ for $k=3$ and $0.57\pm 0.45$ for $k=5$ while gradients themselves have magnitudes of $150-250$.

\section*{Remarks on timings}

Concerning the speed of the approach above it may be noticed that of course the computations required for VMAF are significantly heavier than for simple loss functions such as, e.g., $L_{1}, L_{2}$. However, in many realistic applications of this loss function the timing has but a small importance. For example, in the case of constructing a pre-processing filter by means of training (see the next section) we just can use the loss offline to train the filter, and then the image or video stream can be processed with the filter in real time. Nevertheless, it may be of interest to provide time measurements for several typical loss functions compared to our implementation; they are presented in the following table.
\begin{table}[htbp]
\caption{Inference time for several loss functions; all computations were done on GPU. A pair of HD image (reference and distorted) or a single reference image (for BRISQUE metric) was feeded to the loss function $30$ times and the average time and std were measured.}
\begin{center}
\begin{tabular}{rr}
\toprule
Loss & time, ms \\
\midrule
PSNR & $0.262\pm 0.007$ \\
SSIM & $17.8 \pm 1.5$ \\
MS-SSIM & $24.63 \pm 0.34$ \\
BRISQUE & $20.08 \pm 0.06$ \\
VMAF-torch &  $30.4 \pm 0.2$ \\
LPIPS & $32.05 \pm 0.48$ \\
LPIPS-VGG & $521.6 \pm 2.8$\\
\bottomrule
\end{tabular}
\end{center}
\label{tab:gaussian_laplace_5}
\end{table}
The results show that the speed of VMAF-torch is superior compared to such a popular loss function as LPIPS and quite comparable to MS-SSIM. As was said above, simple losses like PSNR are noticeably faster.

\section*{An application: Training a preprocessing filter}

\begin{figure*}
	\centering
	\begin{subfigure}[b]{0.8\textwidth}
		\centering
		\includegraphics[width=0.8\linewidth]{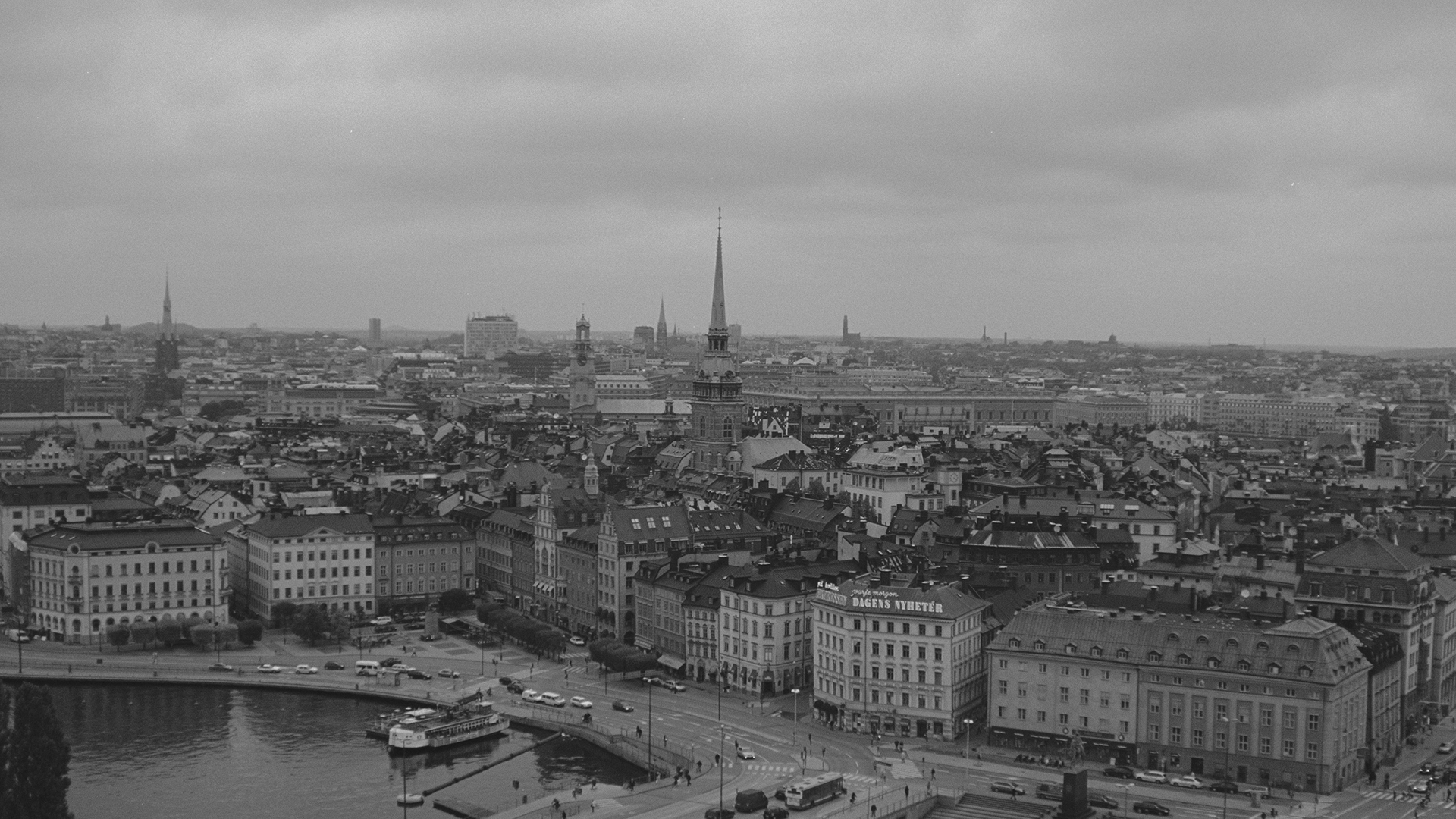}
		\caption{Reference}
		\label{fig:a}
	\end{subfigure}
	\hfill
	\begin{subfigure}[b]{0.8\textwidth}
		\centering
		\includegraphics[width=0.8\textwidth]{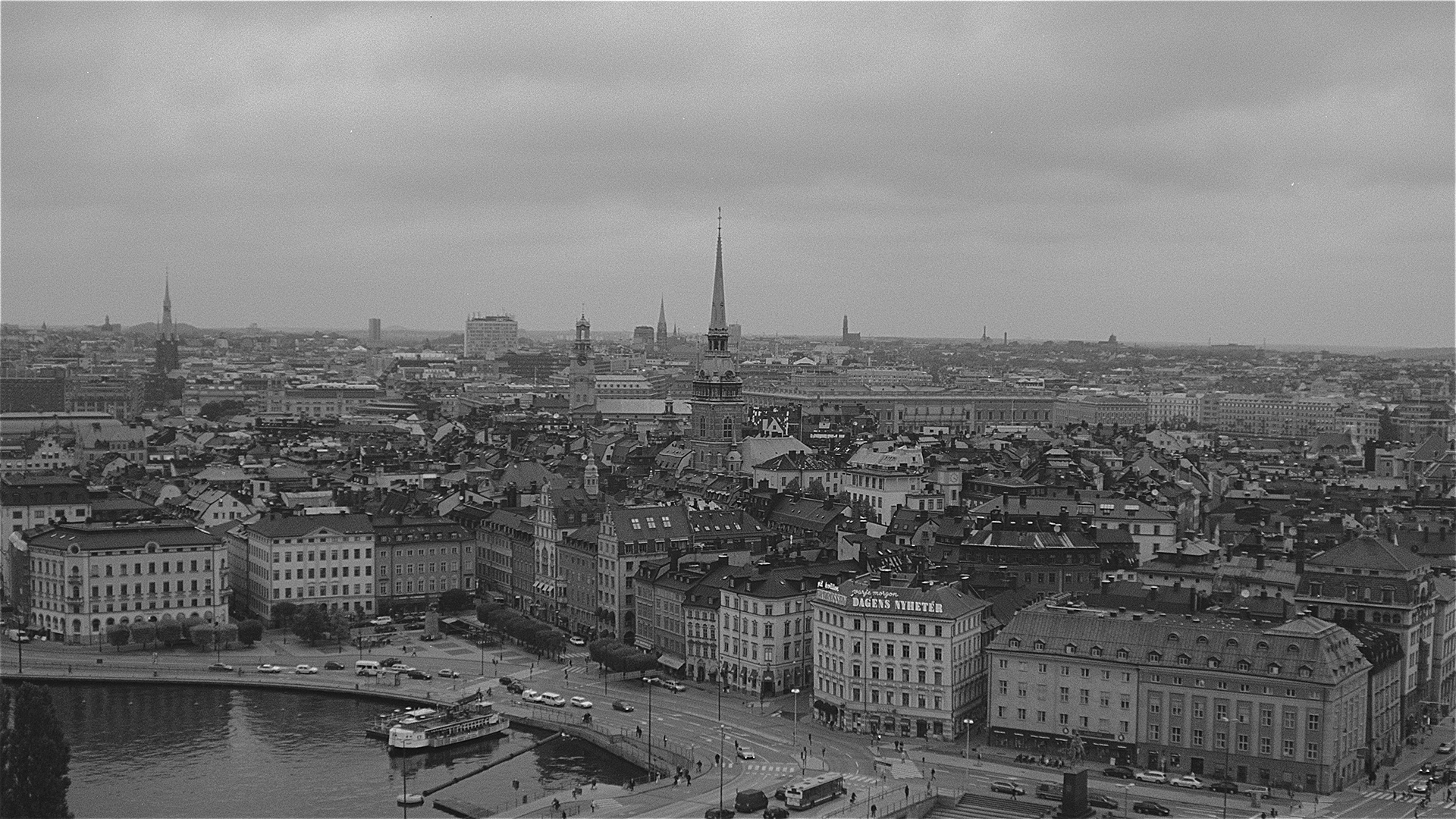}
		\caption{Unsharp masking filter $7\times 7$, $\alpha=1$.}
		\label{fig:b}
	\end{subfigure}
	\hfill
	\begin{subfigure}[b]{0.8\linewidth}
		\centering
		\includegraphics[width=0.8\linewidth]{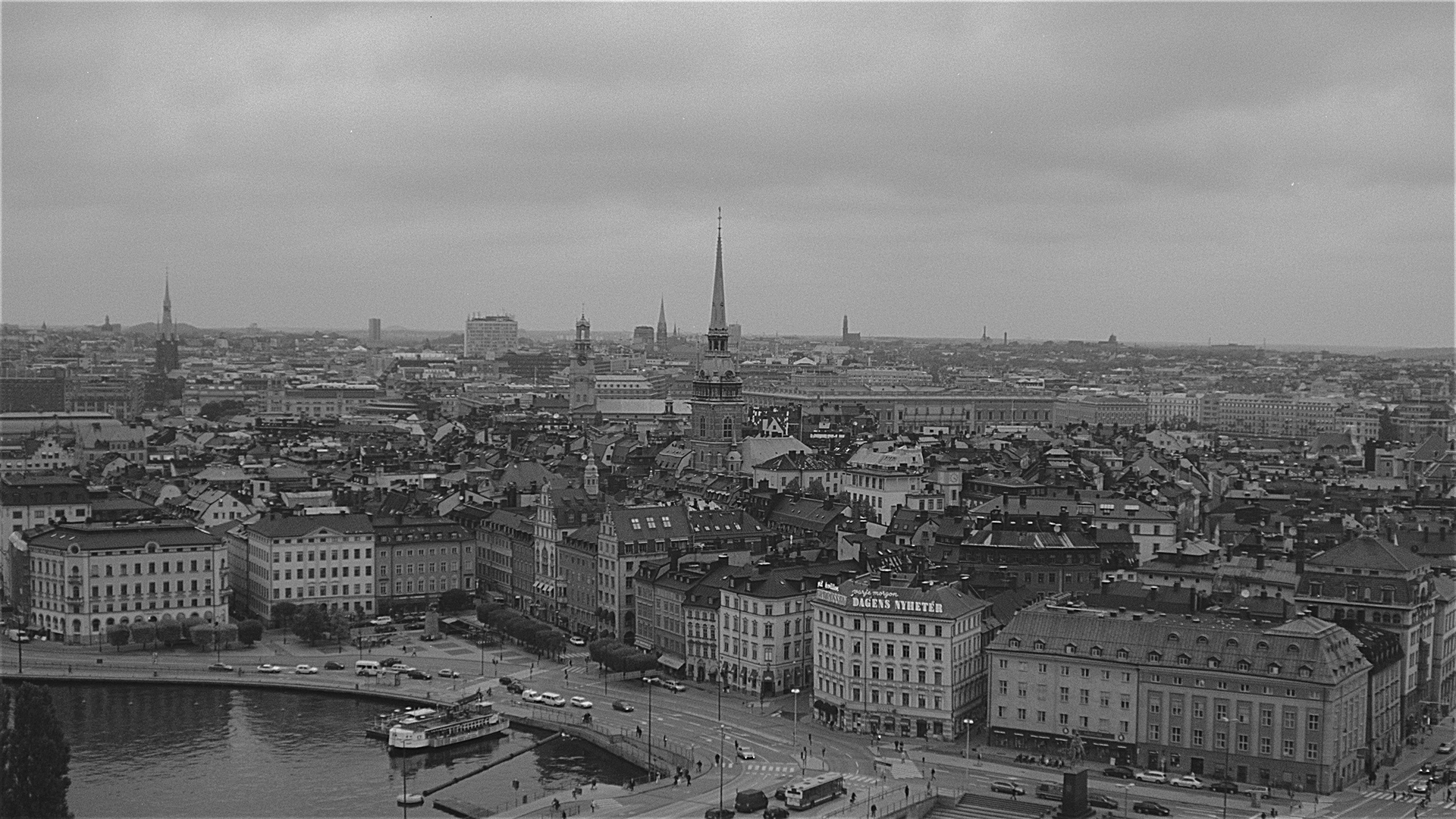}
		\caption{Optimal filter, $7\times 7$, $\alpha=0.5$.}
		\label{fig:c}
	\end{subfigure}
	\caption{Visual comparison of unsharp masking filter (\ref{fig:b}) with the optimal filter (\ref{fig:c}) constructed as described in the main text and the reference image (\ref{fig:a}). The image represents a frame extracted from the publicly available Netflix data set\cite{netflix}.}
	\label{fig:visual_comparison}
\end{figure*}

To assess the applicability of VMAF as a loss function we perform a simple optimization procedure: inspired by unsharp masking filter we attempt to train a single convolutional filter. The unsharp masking filter is a widespread image high-pass filter \cite{adaptive_unsharp} that is used to increase the sharpness of image; it is known to increase VMAF score \cite{siniukov2021hacking}. The unsharp masking filter can be expressed by
$$
U=I+\alpha(I-G),
$$ 
where $I$ is the identity filter (a matrix with $1$ at the center and $0$ everywhere else), $G$ is a gaussian filter and $\alpha$ is a parameter acting as an amplification/attenuation coefficient.
Unsharp masking can also be viewed as a single convolution of small size applied to the luma component of the image. We train a convolutional filter of size $7\times7$ on luma data in the same way as the unsharp masking filter is usually applied. 
Given a batch of images $\{R_i\}_{i=1}^n$ we optimize 
$$
L(W)=\sum_{i=1}^{n}\operatorname{VMAF(R_i, R_i \ast W)}
$$
with respect to the filter coefficients $w_{ij}$ using stochastic gradient descent with learning rate $1\times 10^{-5}$. The weights are initialized with identity filter weights. An additional restriction 
$$
\sum_{ij} w_{ij}=1
$$ 
is applied to keep the average scale for brightness of the image; this condition is also satisfied by unsharp masking filter $U$. To ensure this we normalize the kernel by dividing the elements by their sum at each training step, this can be thought of as a form of projected gradient descent; the details of this procedure will be described elsewhere.
We disable clipping VMAF into $[0,100]$ range since we already start with VMAF scores close to $100$ and the clipping operation zeroes the gradients. We perform early stopping since during training the magnitude of VMAF grows to the infinity, which can be explained by the fact that VMAF score is obtained by SVM regression. This situation, however, can be presumably improved.
The resulting filter $W^*$ is circularly symmetric up to certain precision, while no restriction on symmetry was applied. The results of image processing with unsharp masking and optimal filters are demonstrated in Fig. \ref{fig:visual_comparison}.

\begin{figure}
	\includegraphics[width=1.0\linewidth]{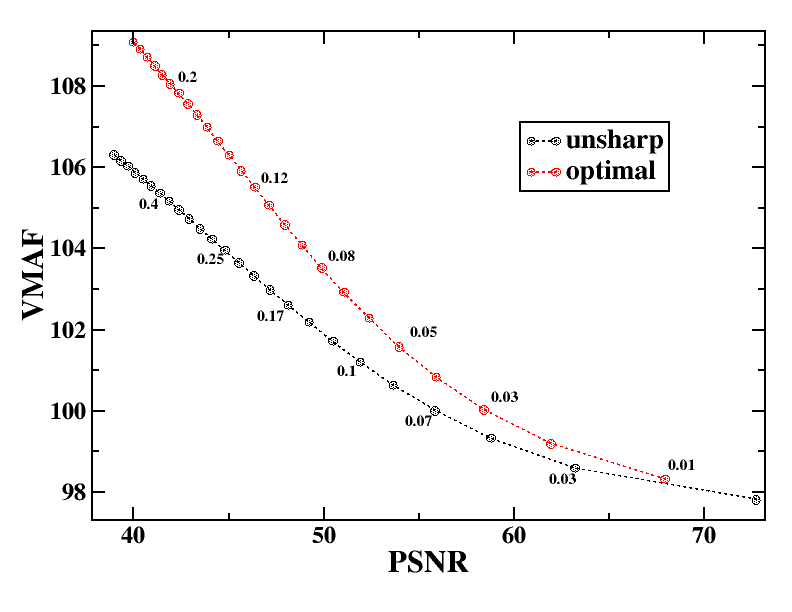}
	\caption{VMAF vs PSNR trade-off for the optimal filter. The computations were done on frames from Netflix public dataset after applying our filter and unsharp masking filter with various values of $\alpha$ parameter (shown next to the points). Note that for $\alpha\to 0$ VMAF score converges to $\sim 97.4$ instead of $100$; this occurs when the Motion feature of VMAF is equal to zero. Both filters have the size $7\times 7$.}
	\label{fig:psnr_vmaf}
\end{figure}

To assess the performance of our filter with respect to the unsharp masking it is not enough to look at VMAF value alone because the growth of the amplification coefficient $\alpha$ in the unsharp masking results in increasing VMAF and lowers PSNR. For the convenience of representation we transform our filter to the form similar to unsharp masking filter $W^{*}=I+\hat W$ and introduce $\alpha$ parameter $W^{*}_{\alpha}:=I+\alpha \hat W$ to make the form of the filter resembling the unsharp masking filter. Increase in $\alpha$ leads to the increase in VMAF and the decrease in PSNR analogous to unsharp masking filter.
The comparison of our optimal learnt filter with the unsharp masking filter for various amplification magnitudes $\alpha$ is provided in Fig. \ref{fig:psnr_vmaf}. 
It is clearly seen that in a wide range of PSNR values the optimal filter yields better image quality in terms of VMAF.

\begin{figure}
	\includegraphics[width=1.0\linewidth]{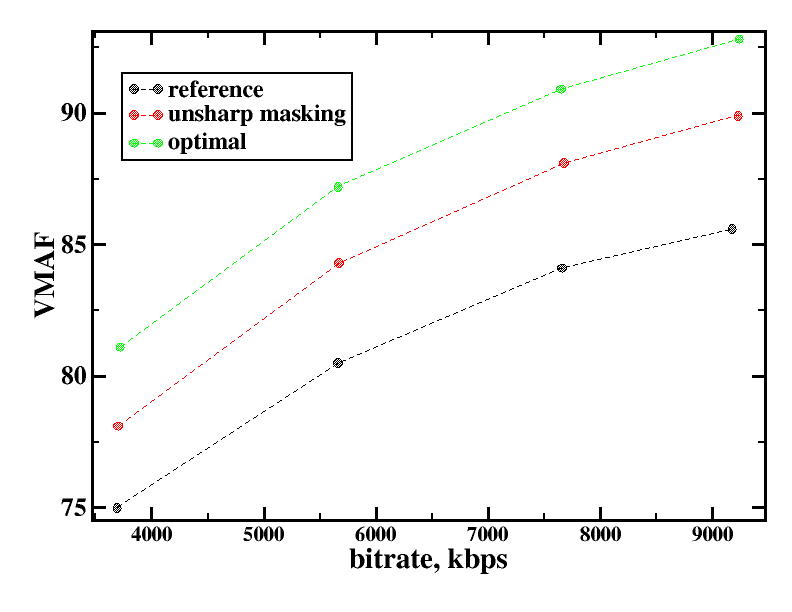}
	\caption{VMAF RD curves were obtained using a synthetic stream presenting a video game. The measurement was done on four target bitrates $4000$, $6000$, $8000$, $9500$ kbps. For the unsharp masking filter $\alpha=0.5$; for the optimal filter $\alpha=0.25$.}
	\label{fig:rd_curves}
\end{figure}

These results were confirmed using HEVC video codec\footnote{We used the proprietary Huawei hw265 video codec for tests. A more extended study and results with open source video codecs are in preparation and will be published elsewhere.} and are shown in Fig. \ref{fig:rd_curves} to compress the streams processed with various filters.
They show that the filter obtained by means of stochastic gradient descent (SGD) method using our implementation of VMAF as a cost function provides better performance compared to the unsharp masking in a range of bitrates. 

\section*{Discussion}

The proposed implementation raises some questions. In the literature one can find claims that VMAF is not differentiable (see, e.g., \cite{ProxVQM} and \cite{venkataramanan2023transform}). Surprisingly, we were not able to find precise clarifications of these claims. This clarification may be of importance as depending on how to understand the differentiability some objective functions can be applicable for training ML models. 
On the one hand, from the optimization point of view perspective, these claims may merely state that the existing C implementation does not, as we stressed in the Introduction, implement functions that compute precise gradients, which is correct. On the other hand, these claims may imply that VMAF as a function is not differentiable in the mathematical sense. We would like to provide a more detailed discussion concerning these claims. 

In the mathematical sense a function is considered differentiable if its derivative exists at each point of its domain. By this definition functions commonly used in machine learning such as RELU or $L_1$ loss turn out to be non-differentiable, since there exist points in their domain in which the derivatives do not exist, e.g. for RELU it is $0$. From this point of view VMAF is also a non-differentiable function because it is constructed as a composition of functions some of which are not differentiable. However, it turns out that even in this case one still can use such functions. For example, instead of the non-existing derivative, left or right derivatives can be used and in fact this is how frameworks like PyTorch deal with the problem of non-differentiability\footnote{https://pytorch.org/docs/stable/notes/autograd.html\#gradients-for-non-differentiable-functions.}. It is this way that a wide variety of functions that are {\it formally non-differentiable} can be used in the machine learning field.

Another worth mentioning, though absolutely {\it ad hoc} meaning of non-differentiable is sometimes used in the literature (e.g. in \cite{reichDifferentiableJPEGDevil2024} section 4): a function is called non-differentiable if all  its first derivatives are equal to zero. It is clear that this definition is relevant to ML algorithms. For such functions the learnable parameters can not be updated using methods such as SGD. 
For example the floor function $f(x) = \lfloor x \rfloor$ has both the left and the right derivative equal to zero on $R$. 
If an objective function is a composition that includes the floor function, then by the chain rule the gradient of the objective function will also be zero. 
Generally speaking, this effect is mainly caused by various types of quantization such as floor, ceiling, rounding to integers, e.g. in \cite{oordNeuralDiscreteRepresentation2018}, \cite{jacobQuantizationTrainingNeural2017}.
Several quite evident methods have been proposed to solve this problem: straight-through estimation (replacing the zero gradients with $1$)\cite{bengioEstimatingPropagatingGradients2013}, \cite{oordNeuralDiscreteRepresentation2018}, \cite{reichDifferentiableJPEGDevil2024}, changing the function with zero derivatives to a similar function with non-zero derivatives \cite{reichDifferentiableJPEGDevil2024}.
Note that such zero-gradient functions are not present in VMAF and we mention it here just for completeness. It can be argued that rounding is used in VMAF since all functions in the default version of VMAF in libvmaf are performed using integers, however a version of VMAF that uses floating point numbers is also available\footnote{e.g. for VIF: integer version \cite{VMAF_integer}, float version \cite{VMAF_float}. } and in it explicit rounding is not performed.

A weaker version of the problem described above may be caused by functions that have derivative equal to zero on some part of their domain, e.g. RELU has the derivative equal to zero on $(-\infty, 0)$.
In this case for some input vectors some elements of the gradient vector may be equal to zero.
In the case of neural networks this is mostly caused by activation functions and may lead to a part of network parameters not being updated by SGD. This is known as dying neuron or dying RELU problem \cite{Lu_2020}.
This problem does affect VMAF in some cases since VMAF internally uses clipping and RELU functions. 
The final VMAF score is known to be clipped to $[0,100]$ range, which leads to zero derivatives for images that have (unclipped) VMAF score of $>100$ or $<0$. However one can easily avoid this by simply not performing this final clipping operation, which we also do in our experiments. Clipping is also used in VMAF internally, e.g., in the equations (2) and (\ref{vmaf_neg_clipping}). These functions can be removed or replaced: e.g., RELU with Leaky RELU \cite{Maas2013RectifierNI} and clipping with soft clipping \cite{reichDifferentiableJPEGDevil2024}. This however will affect the final VMAF score and since our goal is to reproduce VMAF as closely as possible we do not change any internal functions.
In spite of this in all our experiments we do not observe zero gradients. It is possible that for some exotic inputs the gradient will be equal to zero, but we believe that this is not quite probable in practice.

One can note that the definition of VIF metric contains mutual information and generally speaking it is impossible to talk about the derivative wrt. a random variable even though such a definition can be consistently constructed. Probably, in this sense one can say that the mutual information and consequently, VIF and VMAF are non-differentiable. However, for {\it ad hoc} purposes this is not necessary: we can alter the derivative definition so as to obtain the consistently working algorithm. First of all, VIF model implies the gaussian channel model for HVS (human visual system) as well as for the image distortion. This model introduces a set of parameters, e.g., parameters $g_{i}$ in (\ref{vif}) and it seems easy to differentiate (\ref{vif}) wrt. these parameters. Moreover, these parameters may depend on other, hidden parameters, e.g., some filter coefficients $\mathbf{w}$ as we demonstrated in the previous section, so in fact we will have a composition of functions containing $g_{i}(\mathbf{w})$ and would be able to differentiate VIF both wrt. $g_{i}$ and $\mathbf{w}$. It may be worth noting here that some works also try to establish the computability of the gradient of the mutual information wrt. parameters, presumably, in a much more rigorous way \cite{mi_gradient}. Secondly, the implementation of a cost function in PyTorch and sufficiently good behavior of its gradients in the experiments described above may imply its differentiability in {\it this restricted sense}. Thus, in this sense,  we also can roughly say VMAF can be considered as differentiable and what is more important used in gradient descent tasks. This again was confirmed numerically in the computational experiments described above.

For the purposes of gradient descent related algorithms there are attempts to train a convolutional neural network to predict the VMAF score for images \cite{ProxIQA} and video \cite{ProxVQM}. In \cite{ProxIQA} the network is used to optimize a neural net for image compression. The difficulty of this approach may be that the net is not guaranteed to produce the output close to VMAF on input that differs from the training data. Indeed, the authors of \cite{ProxIQA} have to re-train the net continually together with the compression net. 
The proposed implementation of VMAF algorithm, on the other hand, enables us to study the properties of VMAF directly and use this metric as a cost function for various optimization tasks related to compression. A piece of evidence for that was provided above; more detailed results on application of this approach are in preparation and will be published elsewhere.

Our implementation of VMAF reproduces values obtained with the standard implementation with significant precision $\lesssim 10^{-2}$.  We believe that this implementation can be beneficial to image/video quality and compression communities due to its possible use for training neural networks for tasks such as compression, image enhancement and others\footnote{We plan to release this code as an opensource software which can not be fulfilled immediately for security procedures.}. The validity of this implementation is confirmed by the results of the learning procedure and results obtained in application to various state-of-the-art video codecs.

\bibliographystyle{IEEEtran}

\bibliography{IEEE_IQA_and_metrics,IEEE_ML,bibtex/bib/IEEE_Image_processing}

\section*{Acknowledgment}

The authors are grateful to their colleagues in Media Technology Lab Alexey Leonenko, Vladimir Korviakov, and Denis Parkhomenko for helpful discussions.

\end{document}